\documentclass[runningheads]{llncs}
\usepackage{graphicx}

\usepackage{tikz}
\usepackage{comment}
\usepackage{amsmath,amssymb} 
\usepackage{color}

\usepackage[accsupp]{axessibility}  

\usepackage{graphicx}
\usepackage{amsmath}
\usepackage{amssymb}
\usepackage{float}
\usepackage{tabularx}
\usepackage{algorithm}
\usepackage{algpseudocode}
\usepackage{enumerate}
\usepackage{xspace}

\usepackage{enumitem,kantlipsum}
\usepackage[mathcal]{eucal}
\usepackage{graphicx}
\usepackage{amsmath}
\usepackage{amssymb}
\usepackage{float}
\usepackage{tabularx}
\usepackage{algorithm}
\usepackage{algpseudocode}
\usepackage{enumerate}
\usepackage{wrapfig}
\usepackage{makecell}

\usepackage{enumitem,kantlipsum}

\RequirePackage{multirow}  
\RequirePackage{array}	   
\RequirePackage{booktabs}  
\RequirePackage{tabularx}  
\RequirePackage{longtable} 
\RequirePackage{tabu}      
\RequirePackage{threeparttable} 
\usepackage[pagebackref,breaklinks,colorlinks]{hyperref}
\usepackage[capitalize]{cleveref}
\crefname{section}{Sec.}{Secs.}
\Crefname{section}{Section}{Sections}
\Crefname{table}{Table}{Tables}
\crefname{table}{Tab.}{Tabs.}

\newcommand{\ourmethod}{{MoQuad}\xspace}

\makeatletter
\DeclareRobustCommand\onedot{\futurelet\@let@token\@onedot}
\def\@onedot{\ifx\@let@token.\else.\null\fi\xspace}

\def\eg{\emph{e.g}\onedot} 
\def\ie{\emph{i.e}\onedot} 
 
\def\etc{\emph{etc}\onedot}

\makeatother

\newcommand\mypara[1]{\vspace{1mm}\noindent\textbf{#1}}

\begin{document}
\pagestyle{headings}
\mainmatter
\def\ECCVSubNumber{4}  

\makeatletter
\def\blfootnote{\xdef\@thefnmark{}\@footnotetext}
\makeatother

\title{MoQuad: Motion-focused Quadruple Construction for Video Contrastive Learning}

\titlerunning{MoQuad}
%
\author{Yuan Liu\inst{1\,}$^{\dagger}$ \and
Jiacheng Chen\inst{2\,}$^{\dagger}$ \and
Hao Wu\inst{3}}

\authorrunning{Y. Liu et al.}
%
\institute{The University of Hong Kong \and Simon Fraser University \and Bytedance Inc
}


\maketitle
\begin{abstract}
Learning effective motion features is an essential pursuit of video representation learning. This paper presents a simple yet effective sample construction strategy to boost the learning of motion features in video contrastive learning. The proposed method, dubbed \textbf{Mo}tion-focused  \textbf{Quad}ruple Construction (\ourmethod), augments the instance discrimination by meticulously disturbing the appearance and motion of both the positive and negative samples to create a quadruple for each video instance, such that the model is encouraged to exploit motion information. Unlike recent approaches that create extra auxiliary tasks for learning motion features or apply explicit temporal modelling, our method keeps the simple and clean contrastive learning paradigm (\ie, SimCLR) without multi-task learning or extra modelling. In addition, we design two extra training strategies by analyzing initial \ourmethod experiments. By simply applying \ourmethod to SimCLR, extensive experiments show that we achieve superior performance on downstream tasks compared to the state of the arts. Notably, on the UCF-101 action recognition task, we achieve 93.7\% accuracy after pre-training the model on Kinetics-400 for only 200 epochs, surpassing various previous methods. 
\end{abstract}
\blfootnote{\noindent $^{\dagger}$Work done when interned at Bytedance Inc.}
\section{Introduction}
\label{sec:intro}
The recent progress of self-supervised visual representation learning has provided exciting opportunities for exploiting the huge amount of unlabelled web images and videos to train extremely powerful neural networks. In the image domain, seminal works based on contrastive learning (\eg, SimCLR~\cite{simclr}, MoCo~\cite{moco}) have largely closed the performance gap between self-supervised learning methods and the supervised learning counterparts across various downstream tasks. However, the progress in the video domain lags behind as videos contain rich motion information, making the learning of video representations more challenging. 

Contrastive learning is still an effective framework for video self-supervised learning~\cite{Feichtenhofer2021ALS,CVRL}, but to further encourage the learning of motion-oriented features, extra designs are usually needed. For example, RspNet \cite{rspnet}, ASCNet \cite{ascnet} and Pace \cite{pace_prediction} propose an extra motion branch and conduct multi-task learning, while DPC \cite{DPC}, MemDPC \cite{DPC} and VideoMoCo \cite{videomoco} apply explicit temporal modeling for better modeling motion information.

\begin{figure}[t]
\centering
\setlength{\abovecaptionskip}{0.cm}
\includegraphics[width=1.0\textwidth]{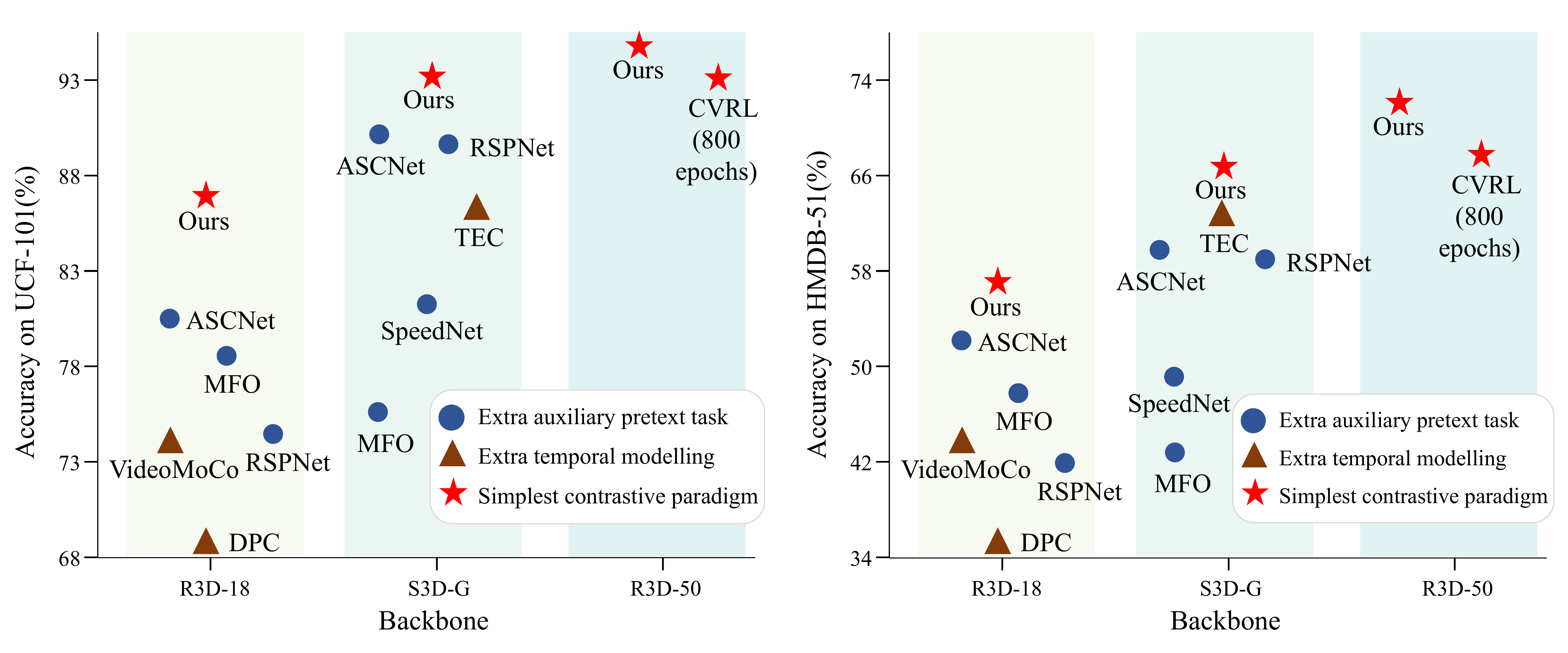}
\vspace{-0.3cm}
\caption{
 Performance of \ourmethod on UCF-101 and HMDB-51 action recognition (transfer learning pre-trained on Kinetics-400~\cite{k400} for only 200 epochs) with different video backbones, compared to other video self-supervised learning methods. The full results corresponding to this figure are available in Table~\ref{tab:big_table}. 
}
\label{fig:tier_1}
\vspace{-0.5cm}
\end{figure}

This paper aims to keep the simplest video contrastive learning paradigm (\ie, single-task, no extra modelling), but significantly improve the learning of motion features. Our main contribution is a carefully designed sample construction strategy that enriches the vanilla instance discrimination and enforces the model to capture effective motion information to finish the discrimination task. 

Concretely, for each anchor clip, the instance discrimination~\cite{Feichtenhofer2021ALS} task requires the model to pull clips from the same video instance (positive sample) closer while pushing clips from different instances (negative sample) apart. Our method then builds upon three progressive analyses: 
\begin{enumerate}[label=(\arabic*),leftmargin=*,topsep=0pt,itemsep=0pt,noitemsep]
\item Positive clip pairs from the same video instance share similar appearance and motion features, but ConvNets tend to excessively exploit the appearance clues (\eg, background) to hack the instance discrimination task, thus neglecting the motion features~\cite{BE}. To prevent the model from overly relying on the appearance clues, it is necessary to disturb the appearance of the positive sample.
\item However, compared to the negative samples from other videos, the appearance disturbed positive sample still shares higher similarity with the anchor in terms of the appearance, thus still leaving space for the model to bypass the motion feature. To alleviate this, we argue that an intra-video negative sample is further needed, which has exactly the same appearance information as the anchor, but with the motion feature being disturbed.
\item To distinguish the appearance-disturbed positive sample from the motion-disturbed intra-video negative sample, there exist two ways: learning effective motion features or hacking the bias introduced by the appearance disturbing operation. To completely block the latter option, we propose to further include one more intra-video negative sample by disturbing both the appearance and motion information. 
\end{enumerate}
Combining these three arguments, our method, \textbf{Mo}tion-focused \textbf{Quad}ruple Construction(\ourmethod),
creates a quadruple for each video instance consisting of 1) an anchor, 2) an appearance-disturbed positive sample, 3) a motion-disturbed intra-video negative sample, and 4) a motion-and-appearance-disturbed intra-video negative sample, and follows the standard contrastive learning paradigm to train the model. We also conduct detailed analyses to determine the appropriate disturbing operations for motion and appearance.

By simply applying \ourmethod to SimCLR, experiments show that we considerably improve the transfer learning performance. In addition, by analyzing the initial \ourmethod experiments, we further derive two training strategies that bring consistent improvements: 1) a warm-up strategy that warms up the model with an appearance-focused instance discrimination task, such that the model can concentrate more on learning motion features when training with \ourmethod and being less distracted by appearance information; and 2) a hard negative sample mining strategy that pushes the model to better discriminate videos with potentially similar motion types.

Without any multi-tasking learning or extra temporal modelling, \ourmethod surpasses state-of-the-art video self-supervised learning methods on various downstream tasks, including action recognition and video retrieval, with shorter pre-training schedules (See Fig.~\ref{fig:tier_1}). Extensive ablation studies and analyses further justify the effectiveness of different components of our method.

\section{Related Work}
\mypara{Self-supervised image representation learning}
Self-supervised image representation learning has become a hot topic in recent years. Previous studies focus on creating pretext tasks explicitly, such as predicting the rotation angle of an image \cite{rotation}, solving a jigsaw puzzle task \cite{jigsaw} and solving a relative patch predicting task \cite{color}. Recently, some works optimize clustering and representation learning jointly \cite{labelling,ExplainingDN,UnsupervisedPO}, or learning image visual representation by discriminating instances from each other through contrastive learning. \cite{simclr} pulls two augmented crops from the same images together while pushing crops from different images apart, using InfoNCE \cite{InfoNCE}. \cite{non-parametric} proposes to use a memory bank to store all image representations. To keep the representations consistent, \cite{moco} proposes MoCo, which stores image embeddings from a momentum update encoder in a queue and achieves superior performance. The above-mentioned methods usually rely on a large number of negative samples. Meanwhile, BYOL \cite{BYLO} and Siamese \cite{siamese} learn meaningful representations by only maximizing the similarity of two augmented positive samples without using any negative samples.

\mypara{Self-supervised video representation learning} Similar to image- representation learning approaches, some video representation learning approaches create pretext tasks \cite{Agrawal2015LearningTS,Diba2019DynamoNetDA,Epstein2020OopsPU,Isola2015LearningVG,Jayaraman2015LearningIR,Lai2020MASTAM,kaihu}. Recently, contrastive learning becomes the mainstream in self-supervised video representation learning \cite{CVRL,DPC,MemoryDPC,coclr}. For example, CVRL \cite{CVRL} pulls two augmented clips from the same video together while pushing clips from different videos apart. Since the motion information is very important in videos, some approaches, \eg \cite{BE}, focus on extracting the motion of videos by preventing the model from focusing on the appearance. To extract both the motion and appearance information of videos, \cite{rspnet,pace_prediction,ascnet} propose to create two branches and achieve satisfactory results.
\section{Method}
\label{sec:method}

Our method consists of 1) the \ourmethod sample construction strategy, which is the key innovation of the paper (Sec~\ref{subsec:quad}), and 2) two extra training strategies derived by analyzing our initial experiments with \ourmethod (Sec~\ref{subsec:training_strategies}). 

\begin{figure*}[t]
  \centering
  \includegraphics[width=1.0\linewidth]{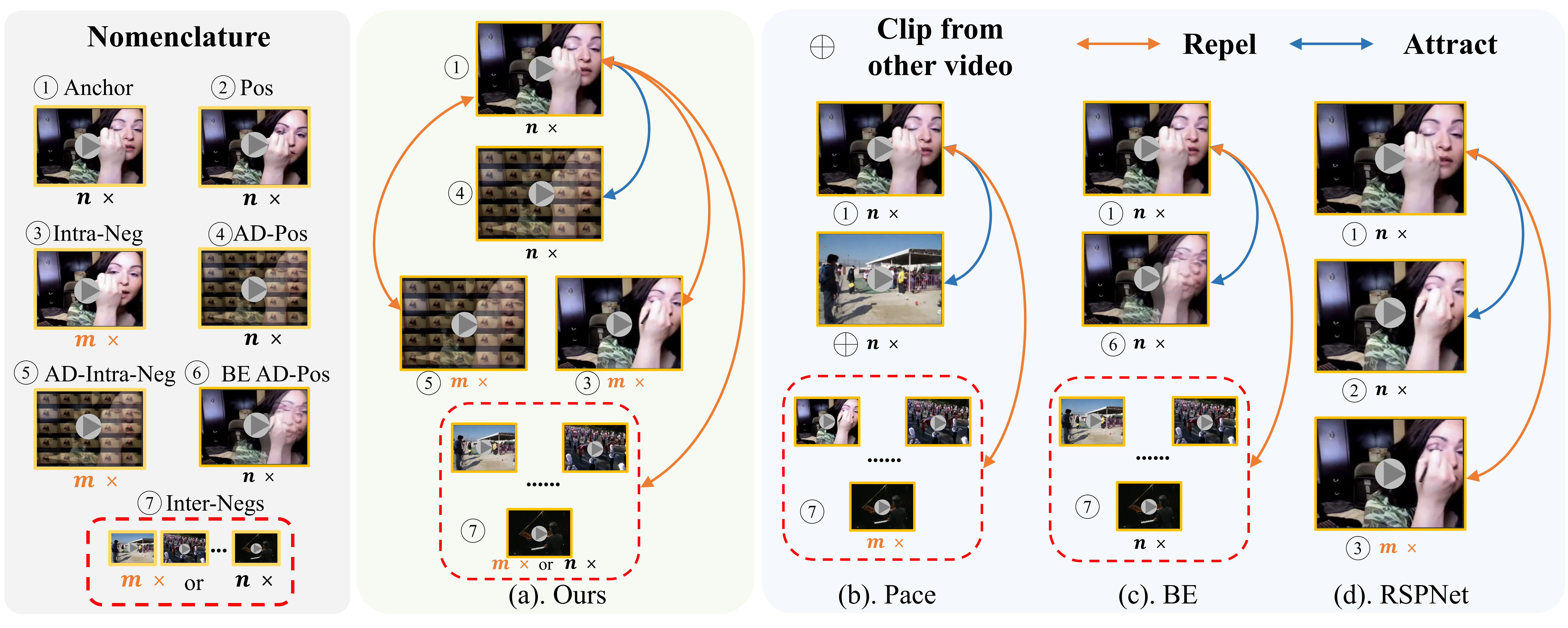}
  \vspace{-0.5cm}
   \caption{\textbf{Comparison between \ourmethod with other motion-focused tasks from previous methods}. (a): \ourmethod, (b): motion task of Pace \cite{pace_prediction}, (c): motion task of BE \cite{BE}, (d): motion task of RSPNet \cite{rspnet}. $n \times$ and $m \times$ denote different video playback speeds. We disturb the appearance of the positive sample to create the appearance-disturbed positive sample and introduce two intra-video negative samples, one with motion disturbed and the other with both motion and appearance disturbed. Inter-Negs are clips from other videos. Please see Sec.~\ref{subsec:quad} for full details.   
   }
   \vspace{-0.2cm}
   \label{fig:main}
\end{figure*}

\subsection{\ourmethod sample construction}
\label{subsec:quad}
\mypara{Definition of the motion-focused quadruple.}
Following the three-step analyses in Sec.\ref{sec:intro}, \ourmethod creates a quadruple from each video instance by carefully disturbing the motion and appearance information, aiming to enforce the learning of effective motion features. Given a video $V_\text{i}$, the quadruple consists of: 
\begin{itemize}[leftmargin=*,topsep=0pt,itemsep=0pt,noitemsep]
    \item An anchor (Anchor): a clip randomly sampled from $V_\text{i}$, denoted as $A_\text{i}$.
    \item An appearance-disturbed positive sample (AD-Pos): a clip randomly sampled from $V_\text{i}$, but with its appearance disturbed, denoted as $\bar{P}_\text{i}$.
    \item An intra-video negative sample (Intra-Neg): a clip randomly sampled from $V_\text{i}$, but with its motion information disturbed, denoted as $N_\text{i}$.
    \item An appearance-disturbed Intra-Neg (AD-Intra-Neg): a clip randomly sampled from $V_\text{i}$, but with both motion and appearance disturbed, denoted as $\bar{N}_\text{i}$.
\end{itemize}
We create the quadruple for each video in a batch, and clips from other videos serve as the inter-video negative samples (Inter-Negs) for the Anchor.

\mypara{Instance discrimination with \ourmethod.}
We pass $A_\text{i}$, $\bar{P}_\text{i}$, $N_\text{i}$, and $\bar{N}_\text{i}$ to the feature extractor \textit{F} and projection head \textit{H} to get the clip features, denoted as $z_\text{iA}$, $\bar{z}_\text{iP}$, $z_\text{iN}$, and $\bar{z}_\text{iN}$, respectively. All the clip features are normalized. The vanilla instance discrimination task is augmented with per-instance quadruple:
\begin{small}
\begin{equation}
\label{eq.motion}
    L_\text{m} = -\sum_{i=0}^{B-1}\log \frac{\exp{(z_\text{iA}\bar{z}_\text{iP} / \tau)}}{\exp{(z_\text{iA}\bar{z}_\text{iP}/\tau)}+\sum{S_\text{Intra}}+\sum{S_\text{Inter}}}
\end{equation}
\end{small}
$\tau$ is the temperature and $B$ is the batch size. We use $S_\text{Intra}$ and $S_\text{Inter}$ to denote $\left\{\exp({z_\text{iA}z_\text{iN}}/\tau), \exp({z_\text{iA}\bar{z}_\text{iN}}/\tau)\right\}$ and $\left\{\exp{(z_\text{iA}z_\text{j}/\tau)}\right\}_{\text{j}\ne \text{i}}$, respectively. 
Note that $z_\text{j}$ includes $z_\text{jA}$, $\bar{z}_\text{jP}$, $z_\text{jN}$ and $\bar{z}_\text{jN}$, where i and j are the indices of two different videos.

\mypara{Criteria for choosing disturbing operations.} 
Determining the concrete disturbing operations is a key step to complete the design of \ourmethod, and we first set up the desiderata for them:
\begin{itemize}[leftmargin=*,topsep=0pt,itemsep=0pt,noitemsep]
    \item Appearance: The appearance disturbing operation is to augment the positive samples (AD-Pos). It should inject sufficiently strong noise into the video clip on the premise of preserving the original motion information, such that the model can be pushed hard to learn motion-focused features. Note that this operation is also applied to AD-Intra-Neg to prevent the model from hacking the bias introduced by the operation (See the analyses in Sec.~\ref{sec:intro}).
    \item Motion: The motion disturbing operation is to augment the negative samples (\ie, Intra-Neg and AD-Intra-Neg). In contrast to the appearance disturbing operation, it should only introduce subtle change to the motion information of a clip to make sure that the discrimination task cannot be trivially solved by the model. Similar arguments are also discussed in~\cite{all}.     
\end{itemize}
With these criteria, we are now ready to choose the disturbing operations.

\begin{figure}[!t]
  \centering
  \setlength{\abovecaptionskip}{0.cm}
  \includegraphics[width=1.0\linewidth, height=0.25\linewidth]{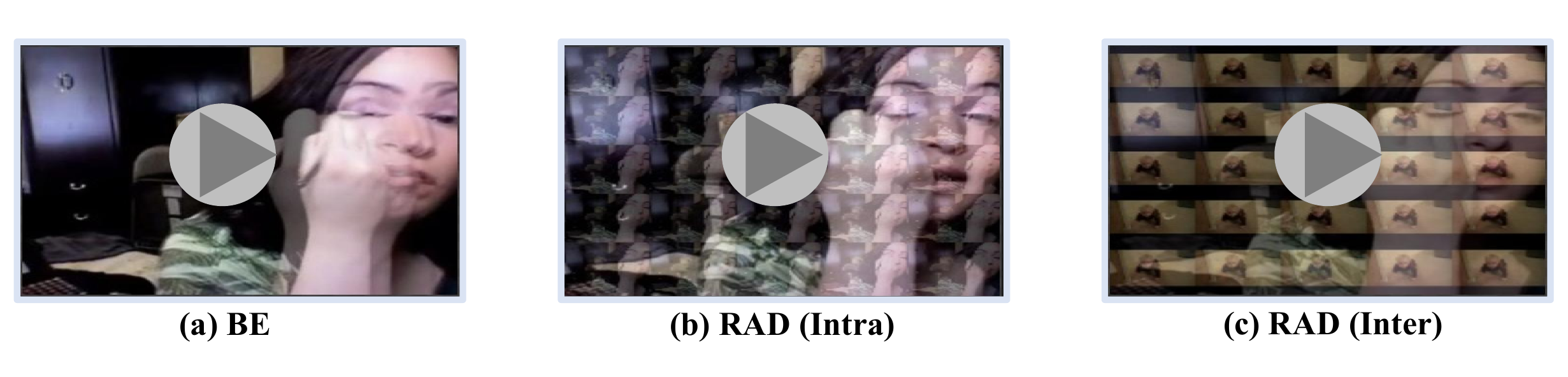}
  \vspace{-0.7cm}
   \caption{\textbf{Appearance disturbing operations}. (a) the operation from BE~\cite{BE}, (b) RAD with the noise image from the current video (intra), and (c) RAD with noise image extracted from another video (inter).}
   \vspace{-0.4cm}
   \label{fig:be}
\end{figure}

\mypara{Appearance disturbing operation.}
Appearance information contains both low and high-frequency information. BE \cite{BE} (Fig.~\ref{fig:be}(a)) provides an off-the-shelf operation by adding to each video frame a noise image randomly picked from the current video. However, this operation still keeps most of the high-frequency information of the original video and is not aggressive enough according to our criteria. To break both the low and high-frequency signals and make the operation stronger, we propose a simple improvement for constructing the noise image. Given a video $V_i$, we create an empty noise image (with the same shape as $V_\text{i}$), denoted as $D$. We then slice $D$ into $k\times k$ windows (we use $k=5$), 
and randomly sample a frame from another video to replace each empty window in $D$. We insert this noise image $D$ to each frame of $V_i$ by a weighted average of 
$
    \bar{V}_\text{i} = (1-\lambda)V_\text{i} + \lambda D
    \label{noise_image}
$.
Fig.~\ref{fig:be}(c) illustrates this operation, and we name it as Repeated Appearance Disturbance (RAD). Note that the repeated frame can be sampled from the current video(Fig.~\ref{fig:be}(b)), but then it only introduces redundant signals to $V_i$ and the RAD (intra) is not as strong as RAD (inter). 

\mypara{Motion disturbing operation.} . 
The motion information in videos can be generally represented as the temporal gradients~\cite{BE}. There are three common approaches to disturb the temporal information of a video in the literature~\cite{rspnet,transformation,shuffle_learn}: 1) change the playback speed (Speed), 2) reverse the frame order (Reverse), or 3) shuffle the frame order (Shuffle). Based on our criteria above, Reverse and Shuffle could change the motion information too much, thus making the negative samples trivial to be discriminated by the model. On the contrary, Speed only modifies the motion information subtly and is a more appropriate choice for \ourmethod.
To be more concrete, given a video $V_\text{i}$, we extract two clips $C_\text{i}^n$ and $C_\text{i}^m$ with dilation rate $n$ and $m$, respectively. $C_\text{i}^m$ and $C_\text{i}^n$ are considered as an Intra-Neg for each other. 
With all operations determined, 
Fig.~\ref{fig:main}(a) is an illustration for the instance discrimination task with \ourmethod quadruples.

\mypara{Relation to previous methods.} 
Pace \cite{pace_prediction}, RspNet \cite{rspnet} and BE \cite{BE} are related previous methods that construct extra samples to improve the learning of motion feature for instance discrimination. All of them demonstrate different properties with \ourmethod, and we make an illustrative comparison in Fig.~\ref{fig:main}.

Pace (Fig.~\ref{fig:main}(b)) enforces the model to pull clips of the same playback speed closer, no matter if these clips are from the same video or not. RspNet (Fig.~\ref{fig:main}(d)) argues that it is inappropriate to compare the speed of clips of different videos, thus only requiring the model to distinguish the playback speed of clips from the same video. Both Pace and RspNet add an extra speed-discrimination task to the contrastive learning framework. However, distinguishing the speed requires the model to focus more on low-level motion features (\eg, the scale of the temporal gradients), and the model could fail to capture high-level motion semantics (\eg motion types) without other negative samples. 
BE (Fig.~\ref{fig:main}(c)) does not use any extra auxiliary task. Instead, it breaks the appearance information of the positive sample to alleviate the negative impact of background bias~\cite{BE}. However, as we analyzed and visualized above, its disturbing operation keeps most of the original appearance clues in the positive sample, and the model might still bypass the learning of effective motion features. 

\ourmethod (Fig.~\ref{fig:main}(a)) does not suffer from the potential issues of these methods. Compared to Pace and RspNet, we consider playback speed as a way to disturb the motion feature of Intra-Neg, instead of making it the only discrimination target (\ie, our Inter-Negs can have the same speed but different motion type as the Anchor). Compared to BE, we employ stronger appearance disturbance by RAD and create extra motion-disturbed negative samples to further prevent the model from bypassing the learning of motion features.

\subsection{Extra training strategies for \ourmethod}
\label{subsec:training_strategies}
In this section, we will introduce the two extra training strategies.
\vspace{-0.5cm}
\subsubsection{Appearance task warm-up.}
When training with \ourmethod, the model exhibits two-stage learning progress: it starts with learning appearance-oriented features and gradually shifts to capture motion-oriented features (visualization in Fig.~\ref{fig:two_stage}(left)), suggesting that the learning of appearance and motion can be better disentangled. Inspired by this discovery, we propose to use an appearance-focused task to warm up the model before training with \ourmethod. We thus borrow the appearance task from RspNet~\cite{rspnet} directly use it as the warm-up task. With the warm-up, \ourmethod focuses on learning motion features from the very beginning of the training and learns better motion features (Fig.~\ref{fig:two_stage}(right)).
Algorithm~\ref{two-stage} details the training schedule with the appearance task warm-up. The appearance task is simply an adapted instance discrimination task:

Given a batch of videos $\textbf{\textit{V}}_B = \left\{ V_\text{i}\right\}_{\text{i}=0}^{B-1}$, for each video $V_\text{i}$ in $\textbf{\textit{V}}_B$, we extract two clips with dilation \textit{n} and \textit{m} respectively, denoted as $C_\text{i}^n$ and $C_\text{i}^m$. We pass $C_\text{i}^n$ and $C_\text{i}^m$ to feature extractor \textit{F} and projection head \textit{H} to get the features, and normalize these features to get the final feature vectors: $z_\text{i}^n$ and $z_\text{i}^m$. The InfoNCE loss \cite{InfoNCE} again pulls clips from the same video together while pushing clips from different videos apart:
\begin{small}
\begin{equation}
\label{app_loss}
    L_\text{a} = - \sum_{i=0}^{B-1}\log \frac{\exp{(z_\text{i}^{n}z_\text{i}^m/\tau)}}{\exp{(z_\text{i}^{n}z_\text{i}^m/\tau)} + \sum_{\text{i}\ne \text{j}, s\in \left\{n, m\right\}}\exp{(z_\text{i}^{n}z_\text{j}^s/\tau)}}
\end{equation}
\end{small}
\vspace{-0.5cm}

\subsubsection{Hard negative sample mining.}
By visualizing the instance discrimination results of \ourmethod, we find that the model sometimes has difficulty in distinguishing the positive samples from hard Inter-Negs. These hard inter-video negative samples turn out to share similar motion semantics as the Anchor (please check the Supplementary for the visualization and more analyses). Based on this observation, we propose a hard negative sample mining strategy to enforce the model to push these hard Inter-Negs farther away from the Anchor.
For all the $\exp({z_\text{iA}z_\text{j}}/\tau)$ in Eq.~\ref{eq.motion}, we first get the top-\textit{K} elements by:
\begin{equation}
    S_\text{Topks} = TopK(\left\{\exp{(z_\text{iA}z_\text{j}/\tau)}\right\}_{\text{i}\ne \text{j}})
\end{equation}
where $K=\beta \times len(\left\{\exp{(z_\text{iA}z_\text{j}/\tau)}\right\}_{\text{i}\ne \text{j}})$ and $\beta$ is the percentage of the total negative samples that are treated as hard negatives. We then assign a large weight, $\alpha(\alpha>1)$, to these hard negative samples, and augment Eq.~\ref{eq.motion} by:
\begin{footnotesize}
\begin{equation}
\label{modified_motion}
    L_\text{m} = -\sum_{i=0}^{B-1}\log \frac{\exp{(z_\text{iA}\bar{z}_\text{iP}/\tau)}}{\exp{(z_\text{iA}\bar{z}_\text{iP}/\tau)}+\alpha \sum{S_\text{Intra}}+\alpha \sum{S_\text{Topks}}+ \sum{S_\text{Easy}}}
\end{equation}
\end{footnotesize}
For notation simplicity, $S_\text{Easy}$ is defined as $S_\text{Easy} = S_\text{Inter} - S_\text{Topks}$ where $-$ is the difference between two sets. 
\begin{algorithm}
\caption{Two-Stage Training Mechanism}
\label{two-stage}
\begin{algorithmic}[1]
\Require video set \textit{V}, feature extractor \textit{F}, projection head \textit{H}, total epochs \textit{E}, current epoch \textit{e} and the percentage of total epochs \textit{p}, used for the appearance task.
\State Initialize $e \leftarrow 0$ \\
// \emph{\textbf{Appearance Task Warmup}}
\While{$e < p \times E$}
\State Sample a batch of videos $V_B = \left\{ V_\text{i}\right\}_{\text{i}=0}^{B-1}$ from 
video sets \textit{V}.
\State For each video in $V_B$, we extract two clips, $C_\text{i}^n$ and $C_\text{i}^m$.
\State Pass these clips to \textit{F} and \textit{H} to get $z_\text{i}^n$ and $z_\text{i}^m$.
\State Use Eq.\ref{app_loss} to finish the instance discrimination task.

\State $e \leftarrow e + 1$
\EndWhile \\
// \emph{\textbf{Instance Discrimination with \ourmethod quadruple}}
\While{$p \times E \leqslant e < E$}
\State Sample a batch of videos $V_B = \left\{ V_\text{i}\right\}_{\text{i}=0}^{B-1}$ from 
video sets \textit{V}.
\State For each video in $V_B$, we create the quadruple: $A_\text{i}$, $\bar{P}_\text{i}$, $N_\text{i}$ and $\bar{N}_\text{i}$
\State Pass $A_\text{i}$, $\bar{P}_\text{i}$, $N_\text{i}$,$\bar{N}_\text{i}$ to \textit{F} and \textit{H} to get $z_\text{iA}$, $\bar{z}_\text{iP}$, $z_\text{iN}$ and $\bar{z}_\text{iN}$.
\State Use Eq.\ref{eq.motion} to finish the instance discrimination task.
\State $e \leftarrow e + 1$
\EndWhile
\end{algorithmic}
\end{algorithm}

\section{Experiments}
\label{sec:exp}
We first describe the experimental setups in Sec.~\ref{subsec:setting}, including the datasets, pre-training details, evaluation settings, \etc. In Sec.~\ref{subsec:compare}, we compare \ourmethod with the state of the arts on two downstream tasks: action recognition and video retrieval. We then provide ablation studies in Sec.~\ref{subsec:ablation} to validate different design choices. In Sec.~\ref{subsec:analyses}, we further present additional analyses to help understand the effectiveness of our method.


\subsection{Experimental settings}
\label{subsec:setting}
\mypara{Datasets.}
Four video action recognition datasets are covered in our experiments: Kinetics-400 (K400)~\cite{k400}, UCF101~\cite{ucf101}, HMDB51~\cite{hmdb51} and Something-Something-V2 (SSv2)~\cite{sthsthv2}. K400 consists of 240K training videos from 400 human action classes, and each video lasts about 10 seconds. UCF101 contains 13,320 YouTube videos from 101 realistic action categories. HMDB51 has 6,849 clips from 51 action classes. SSv2 provides 220,847 videos from 174 classes and the videos contain more complex action information compared to UCF101 and HMDB51.

\mypara{Self-supervised pre-training.} There are various pre-training settings in the literature, we follow the same hyper-parameter choices as CVRL~\cite{CVRL}, including the image resolution ($224 \times 224$) and the number of frames per clip (16). We evaluate our method with three different backbones:  R3D-18~\cite{r3d-18}, S3D-G~\cite{s3d-g}, and R3D-50~\cite{CVRL}, so that we can compare with various previous works that use backbones with different model size. When pre-trained on K400, SSv2, and UCF101, the model is trained for 200, 200, and 800 epochs, respectively, which follows the most common pre-training schedule in the literature~\cite{rspnet,kaihu,CVRL,pace_prediction}. Due to limited computational resources, we conduct ablation studies by pre-training the small R3D-18 backbone on UCF101 by default, unless otherwise specified.
We use LARS~\cite{lars} as our optimizer with a mini-batch of 256, which is consistent with SimCLR~\cite{simclr}. The learning rate is initialized to 3.2, and we use a half-period cosine learning rate decay scheduling strategy~\cite{warmup}. For appearance disturbing, we set $k=5$ and uniformly sample $\lambda$ from $[0.1, 0.5]$. 
For the warmup with appearance task, the appearance task takes up the first 20\% of the training epochs. $\beta$ and $\alpha$ are set to be 0.01 and 1.5 respectively for the hard negative sample mining. 

\mypara{Supervised fine-tuning for action recognition.}
Consistent with previous works~\cite{ascnet,CVRL,DPC}, we fine-tune the pre-trained models on UCF101 or HMDB51 for 100 epochs and report the accuracy of action recognition. We sample 16 frames with dilation of 2 for each clip. The batch size is 32, and we use LARS~\cite{lars} as the optimizer. The learning rate is 0.08, and we employ a half-period cosine learning rate decay strategy.

\mypara{Linear evaluation protocol for action recognition.}
We also evaluate the video representations with the standard linear evaluation protocol~\cite{CVRL}. 
We sample 32 frames for each clip with a temporal stride of 2 from each video. The linear classifier is trained for 100 epochs. 

\mypara{Evaluation details for action recognition.}
For computing the accuracy of action recognition, we follow the common evaluation protocol~\cite{slowfast}, which densely samples 10 clips from each video and employs a 3-crop evaluation. The softmax probabilities of all the 10 clips are averaged for each video to get the final classification results.

\begin{table}[!t]
\centering
\caption{\textbf{Action recognition results on UCF101 and HMDB51.} We report the fine-tuning results of \ourmethod and various baselines under different setups. 
\ourmethod$^{\dag}$ is our method with the two extra training strategies. $^{800}$: pre-trained for 800 epochs, which is longer than the standard training schedule (\ie, 200 epochs).}
\scalebox{1.0}{
\begin{tabular}{lcccccc}
\toprule  
Method&Dataset&Backbone&Frame&Resolution&UCF101& HMDB51\\
\midrule  
CoCLR \cite{coclr}$_{2020}$&UCF101&S3D&32&128&81.4&52.1\\
BE \cite{BE}$_{2021}$&UCF101&R3D-34&16&224&83.4&53.7\\
\ourmethod (Ours)&{UCF101}&{R3D-18}&16&224&80.9&51.0\\
\ourmethod$^{\dag}$ (Ours)&{UCF101}&{R3D-18}&16&224&82.7&53.0\\
\ourmethod$^{\dag}$ (Ours)&{UCF101}&{S3D-G}&16&224&\textbf{87.4}&\textbf{57.3}\\
\hline
Pace \cite{pace_prediction}$_{2020}$&K400& R(2+1)D&16&112&77.1& 36.6\\
MemoryDPC \cite{MemoryDPC}$_{2020}$&K400&R3D-34&40&224&86.1&54.5\\
BE \cite{BE}$_{2021}$&K400&R3D-34&16&224&87.1& 56.2\\
CMD \cite{CMD}$_{2021}$&K400&R3D-26&16&112&83.7&55.2\\
\hline
DPC \cite{DPC}$_{2019}$&K400&R3D-18&40&224&68.2&34.5\\
VideoMoCo \cite{videomoco}$_{2021}$&K400&R3D-18&32&112&74.1& 43.1\\
RSPNet \cite{rspnet}$_{2021}$&K400&R3D-18&16&112&74.3&41.8\\
MFO \cite{MFO}$_{2021}$&K400&R3D-18&16&112&79.1&47.6\\
ASCNet \cite{ascnet}$_{2021}$&K400&R3D-18&16&112&80.5&52.3\\
\ourmethod(Ours)&K400&R3D-18&16&224&85.6&56.2\\
{\ourmethod$^{\dag}$ (Ours)}&{K400}&{R3D-18}&16&224&\textbf{87.3}& \textbf{57.7}\\
\hline
MFO \cite{MFO}$_{2021}$&K400&S3D&16&112&76.5&42.3\\
CoCLR \cite{coclr}$_{2020}$&K400&S3D&32&128&87.9& 54.6\\
SpeedNet \cite{speednet}$_{2020}$&K400&S3D-G&64&224&81.1& 48.8\\
RSPNet \cite{rspnet}$_{2021}$&K400&S3D-G&64&224&89.9& 59.6\\
TEC \cite{TEC}$_{2021}$&K400&S3D-G&32&128&86.9&63.5\\
ASCNet \cite{ascnet}$_{2021}$&K400&S3D-G&64&224&90.8&60.5\\
\ourmethod (Ours)&K400&S3D-G&16&224&91.9& 64.7\\
{\ourmethod$^{\dag}$ (Ours)}&{K400}&{S3D-G}&16&224&\textbf{93.0}& \textbf{65.9}\\
\hline
CVRL \cite{CVRL}$_{2020}^{800}$&K400&R3D-50&16&224&92.9&67.9\\
\ourmethod (Ours)&K400&R3D-50&16&224&93.0&66.9\\
\ourmethod$^{\dag}$ (Ours)&K400&R3D-50&16&224&93.7&68.0\\
{\ourmethod$^{\dag}$ $^{800}$}(Ours)&{K400}&{R3D-50}&16&224&\textbf{94.7}&\textbf{71.5}\\
\bottomrule 
\end{tabular}
}
\label{tab:big_table}
\vspace{-0.8cm}
\end{table}
 
\begin{table}[!t]
\setlength\tabcolsep{4pt}
\centering
\caption{\textbf{Action recognition results on two large-scale datasets.} We further provide linear evaluation results on K400 and SSv2 to demonstrate the effectiveness of our method. Following the two recent works: CVRL~\cite{CVRL} and COP$_{f}$ \cite{kaihu}, 
a R3D-50 backbone is pre-trained on the training split of the corresponding dataset for 200 epochs before conducting the standard linear evaluation. $^{\ast}$: our re-implemented version.}
\begin{tabular}{lccccc}
\toprule  
Method &Backbone&Frame&Batch Size&Resolution&Linear eval.\\
\midrule  
\multicolumn{6}{c}{Pre-trained and evaluated on K400 dataset } \\
\midrule
CVRL \cite{CVRL}$_{2021}$&R3D-50&16&512&224&62.9\\
COP$_{f}$ \cite{kaihu}$_{2021}$&R3D-50&16&512&224&63.4\\
\ourmethod (Ours)&R3D-50&16&256&224&64.4\\
\midrule
\multicolumn{6}{c}{Pre-trained and evaluated on SSv2 dataset } \\
\midrule
CVRL \cite{CVRL}$_{2021^{\ast}}$&R3D-50&16&256&224&31.5\\
COP$_{f}$ \cite{kaihu}$_{2021}$&R3D-50&16&512&224&41.1\\
\ourmethod (Ours)&R3D-50&16&256&224&44.0\\
\bottomrule 
\end{tabular}
\label{ks}
\end{table}

\subsection{Comparison with state of the arts}
\label{subsec:compare}
We compare \ourmethod with state-of-the-art self-supervised learning approaches on action recognition and video retrieval.

\mypara{Evaluation on action recognition.}
As presented in Table~\ref{tab:big_table}, we compare our method with state of the arts under various experimental setups, with UCF101 and HMDB51 as the downstream datasets. Our method always outperforms the previous works with a clear margin when given the same pre-training settings. We note that even without the two extra training strategies, the vanilla \ourmethod is still superior to previous methods.  

As mentioned in Sec~\ref{subsec:setting}, UCF101 and HMDB51 are relatively small-scale datasets. To strengthen the comparison results, we further evaluate our model with the linear evaluation protocol on K400 and SSv2 datasets (Table~\ref{ks}). Our method can still consistently outperform previous works even with smaller effective batch size (\ie, the number of different video instances in a batch).

\begin{table}[!t]
\centering
\caption{\textbf{Evaluation on the video retrieval task}.}
\tabcolsep 4pt
\scalebox{1.0}{
\begin{tabular}{lcccccc}
\toprule  
\multirow{2}*{Method}&\multirow{2}*{Backbone}&\multirow{2}*{Frame}&\multirow{2}*{Resolution}& \multicolumn{3}{c}{Top-\textit{k}}\\
\cmidrule(lr){5-7}
&&&&$k=1$&$k=5$&$k=10$ \\
\midrule
ClipOrder \cite{cliporder}$_{2019}$&R3D-18&16&112&14.1&30.3&40.0 \\
SpeedNet \cite{speednet}$_{2020}$&S3D-G&64&224&13.0&28.1&37.5 \\
MemDPC \cite{MemoryDPC}$_{2020}$&R(2+1)D&40&224&20.2&40.4&52.4 \\
VCP \cite{VCP}$_{2019}$&R3D-18&16&112&18.6&33.6&42.5 \\
Pace \cite{pace_prediction}$_{2020}$&R(2+1)D&16&224&25.6&42.7&51.3\\
CoCLR-RGB \cite{coclr}$_{2020}$&S3D-G&32&128&53.3&69.4&76.6\\
RSPNet \cite{rspnet}$_{2021}$&R3D-18&16&112&41.1&59.4&68.4\\
ASCNet \cite{ascnet}$_{2021}$&R3D-18&16&112&58.9&76.3&82.2\\
{\ourmethod (Ours)}&{R3D-18}&16&224&\textbf{60.8}&\textbf{77.4}&\textbf{83.5}\\
\bottomrule 
\end{tabular}
}
\label{tab:retrieval}
\vspace{-0.6cm}
\end{table}

\mypara{Evaluation on video retrieval.}
We evaluate the learned video representation with nearest neighbour video retrieval. Following previous works \cite{rspnet,pace_prediction}, we sample 10 clips for each video uniformly and pass them to the pre-trained model to get the clip features. Then, we apply average-pooling over the 10 clips to get the video-level feature vector. We use each video in the test split as the query and look for the \textit{k} nearest videos in the training split of UCF101. The feature backbone is R3D-18, and the model is pre-trained on UCF101. The top-\textit{k} retrieval accuracy ($k=1,5,10$) is the evaluation metric. As in Table \ref{tab:retrieval}, we outperform previous methods consistently on all three metrics. 
Note that the gap between our method and recent work, ASCNet~\cite{ascnet}, is small, especially considering ASCNet has a smaller resolution. However, we do outperform ASCNet in Table~\ref{tab:big_table} under the same experimental settings (\ie, resolution 224, pre-trained on K400, and S3D-G backbone). A potential explanation is that fine-tuning could maximize the benefits of \ourmethod.

\subsection{Ablation studies}
\label{subsec:ablation}
This subsection provides ablation studies for different design choices of the paper. As described in Sec.~\ref{subsec:setting}, all ablation experiments use an R3D-18 backbone due to the limited computational resources and a large number of trials.


\begin{table}[t]
\setlength\tabcolsep{4pt}
\caption{\textbf{Ablation studies for components of \ourmethod quadruple}. We pre-train the R3D-18 backbone on UCF101 and K400, respectively. The fine-tuning and linear evaluation results on UCF101 and HMDB51 are reported.}
\centering
\scalebox{1.0}{
\begin{tabular}{ccccccc}
\toprule  
\multirow{2}*{AD-Pos}&\multirow{2}*{Intra-Neg}&\multirow{2}*{AD-Intra-Neg}& \multicolumn{2}{c}{Fine-tuning} & \multicolumn{2}{c}{Linear eval.} \\
\cmidrule(lr){4-5}\cmidrule(lr){6-7}
&&& UCF & HMDB & UCF &HMDB \\
\midrule
\multicolumn{7}{c}{Pre-trained on UCF101 dataset} \\
\midrule
- & - & - &76.0&44.3&68.8&33.6\\
\checkmark & - & - & 78.2 & 47.2 & 70.0& 37.0\\
\checkmark& \checkmark & - & 79.6&49.3&71.5&38.8\\
\checkmark& \checkmark& \checkmark&\textbf{80.9}&\textbf{51.0}&\textbf{73.9}&\textbf{40.3}\\
\midrule
\multicolumn{7}{c}{Pre-trained on K400 dataset } \\
\midrule
- & - & - &80.9&50.2&75.0&44.3\\
\checkmark& - & - &82.1&53.6&76.2&47.4\\
\checkmark& \checkmark& - &84.5&55.0&77.3&48.4\\
\checkmark& \checkmark& \checkmark&\textbf{85.6}&\textbf{56.2}&\textbf{79.2}&\textbf{50.1}\\
\bottomrule 
\end{tabular}}
\label{mpn}
\end{table}

\mypara{The design of quadruple.} We validate the effectiveness of each component of our quadruple in Table~\ref{mpn}.
We use SimCLR \cite{simclr} as the base method and progressively add the elements of the quadruple. Both the fine-tuning and linear evaluation results are reported for UCF101 and HMDB51. Regardless of the pre-training dataset, each element of \ourmethod quadruple improves the performance consistently. We would like to point out that K400 is much larger than UCF101 (see Sec.~\ref{subsec:setting} for dataset stats), indicating that the benefits of our method do not degenerate as the size of the pre-training dataset grows.


\begin{table}[!t]
    \caption{\textbf{Ablation studies for the disturbing operations}. \textbf{(Left)} Different appearance disturbing operations. \textbf{(Right)} Different motion disturbing operations. The models are pre-trained on UCF101 for 800 epochs.}
    \begin{minipage}{0.5\textwidth}
    \label{disturbing}
    \centering
    \begin{tabular}{lcccc}
    \toprule  
    \multirow{2}*{Operations} & \multicolumn{2}{c}{Fine-tuning} & \multicolumn{2}{c}{Linear eval.} \\
    \cmidrule(lr){2-3}\cmidrule(lr){4-5}
    & UCF & HMDB & UCF & HMDB \\
    \midrule
    BE \cite{BE}&79.7&48.5&71.8&39.4\\
    RAD (Intra) &80.1&49.0&72.0&39.4\\
    RAD (Inter) &{80.9}&{51.0}&{73.9}&{40.3}\\
    \bottomrule 
    \end{tabular}
    \end{minipage} %
    \begin{minipage}{0.5\textwidth}
    \centering
    \begin{tabular}{lcccc}
    \toprule  
    \multirow{2}*{Operations} & \multicolumn{2}{c}{Fine-tuning} & \multicolumn{2}{c}{Linear eval.} \\
    \cmidrule(lr){2-3}\cmidrule(lr){4-5}
    & UCF& HMDB & UCF & HMDB \\
    \midrule
    Reverse & 76.3 & 45.4 & 69.8 & 34.7\\
    Shuffle &77.4&46.2&70.3&36.4\\
    Speed &{80.9}&{51.0}&{73.9}&{40.3}\\
    \bottomrule 
    \end{tabular}
    \end{minipage}
\vspace{-0.5cm}
\end{table}

\mypara{The choice of disturbing operations.} We present the results of different motion and appearance disturbing operations in Table~\ref{disturbing}. The left sub-table shows that RAD (inter) is better than the simple trick proposed by~\cite{BE} and RAD (intra), justifying our analysis in Sec.~\ref{subsec:quad}.

As also discussed in Sec.~\ref{subsec:quad}, reversing the frame order (Reverse), shuffling the frames (Shuffle), and changing video playback speed (Speed) are three commonly used tricks to break the motion information of a video. As shown by the comparison in Table~\ref{disturbing} (right), Speed produces the best results, which is consistent with our analyses in Sec.~\ref{subsec:quad}. Reverse and Shuffle could change the motion feature of a video too aggressively, thus making the negative sample easy to be discriminated by the model.

\begin{table}[!t]
\tabcolsep 3.5pt
    \caption{\textbf{Ablation studies for the two extra training strategies.} \textbf{(Left)} The hard negative sample mining strategy. $\beta=0$ and $\alpha=0$ is the best entry in Table~\ref{mpn}(top). \textbf{(Right)} Appearance task warm-up. The entry of $0\%$ is the best entry in the left sub-table. All models are pre-trained on UCF101 for 800 epochs.}
    \begin{minipage}{0.5\textwidth}
    \centering
    \begin{tabular}{cccccc}
    \toprule  
    \multirow{2}*{$\beta$}&\multirow{2}*{$\alpha$} & \multicolumn{2}{c}{Fine-tuning} & \multicolumn{2}{c}{Linear eval.} \\
    \cmidrule(lr){3-4}\cmidrule(lr){5-6}
    && UCF& HMDB & UCF & HMDB \\
    \midrule
    0&0&80.9&51.0&73.9&40.3 \\
    0.01&1.5&\textbf{81.7}&\textbf{52.3}&\textbf{74.9}&41.4 \\
    0.01&2.0&81.5&51.6&74.8&\textbf{41.6} \\
    0.01&3.0&80.9&50.8&74.2&41.0 \\
    0.05&1.5&79.3&49.7&72.7&40.3 \\
    \bottomrule 
    \end{tabular}
    \end{minipage} %
    \begin{minipage}{0.5\textwidth}
    \centering
    \begin{tabular}{ccccc}
    \toprule  
    \multirow{2}*{\makecell[c]{Warmup \\ ratio}} & \multicolumn{2}{c}{Fine-tuning} & \multicolumn{2}{c}{Linear eval.} \\
    \cmidrule(lr){2-3}\cmidrule(lr){4-5}
    & UCF & HMDB & UCF &HMDB \\
    \midrule
    0\%&81.7&52.3&74.9&41.4\\
    10\%&\textbf{82.8}&53.0&75.6&42.0\\
    20\% &82.7&\textbf{53.0}&\textbf{76.6}&\textbf{42.9}\\
    40\%&81.0&50.8&74.2&40.8\\
    60\%&80.0&49.7&72.1&39.9\\
    \bottomrule 
    \end{tabular}
    
    \end{minipage}
    \label{strategy}
\end{table}
\begin{figure}[t]
  \setlength{\abovecaptionskip}{0.cm}
  \centering
  \includegraphics[width=0.7\linewidth]{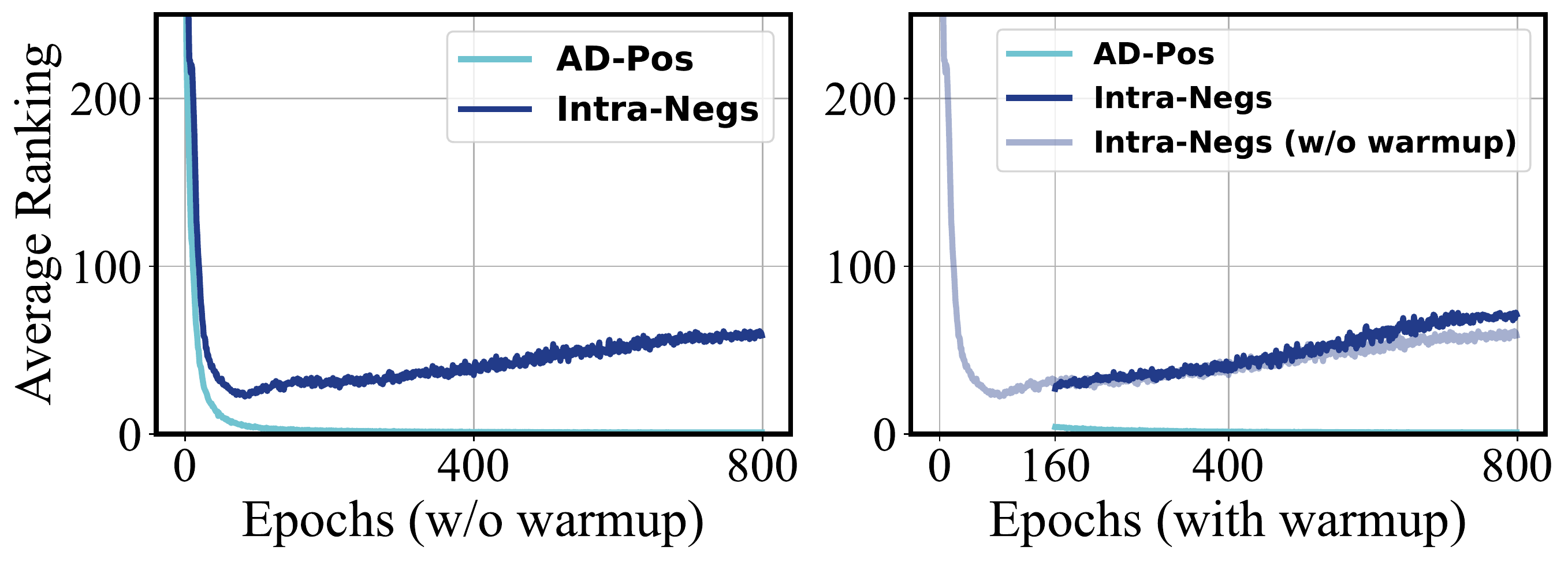}
   \caption{\textbf{Understanding the improvement brought by the warmup of appearance task.} We plot the average rankings of the AD-Pos and the Intra-Negs.
   Intra-Negs includes Intra-Neg and the AD-Intra-Neg. Note that there are no Intra-Negs in the appearance task, thus we start plotting the right figure after the warm-up stage. The warm-up with the appearance task increases the average ranking of Intra-Negs.
   }
   \label{fig:two_stage}
   \vspace{-0.5cm}
\end{figure}

\mypara{Extra training strategies.} 
Table~\ref{tab:big_table} has shown that the two training strategies provide consistent improvements to \ourmethod. We further validate the effectiveness of each strategy and the choice of the hyper-parameters in Table~\ref{strategy}. The hard negative sampling mining and the warm-up with appearance task increase accuracy with proper hyper-parameters. 
To help better understand the warm-up strategy, we plot the ranking of different training samples from our quadruple in the pre-training process in Fig.~\ref{fig:two_stage}. 
As shown in the left figure, the ranking of the Intra-Negs decreases with that of AD-Pos, implying the model first focuses on the appearance of the video. While the right figure shows the warm-up strategy could increase the average ranking of the Intra-Negs during the training process. Note that a higher average ranking of Intra-Negs suggests that the model can better distinguish the subtle differences in motion information injected by the motion disturbing operation.

\subsection{More analyses}
\label{subsec:analyses}
\begin{table}[!t]
\setlength\tabcolsep{4pt}
\centering
\caption{\textbf{Per-category accuracy on SSv2.} We compare \ourmethod with our base method, SimCLR~\cite{simclr}, on fine-grained splits of SSv2 dataset to further demonstrate the capacity of our method to learn better motion features. The categories on the top can mostly be recognized using only the appearance information, while the bottom categories require deeper understanding of the motion in the video.}
\begin{tabular}{lcc}
\toprule  
Video category in SSv2 &SimCLR&\ourmethod\\
\midrule
\multicolumn{3}{c}{Categories that do not need strong temporal information to classify} \\
\midrule
Tear something into two pieces&75.2&83.9\\
Approach something with camera&60.1&88.9\\
Show something behind something&56.3&57.4\\
Plug something into two pieces&57.2&56.3\\
Hold something&27.4&26.3\\
\midrule
\multicolumn{3}{c}{Categories that need heavy temporal information to classify} \\
\midrule
Move something and something closer&41.4&75.2\\
Move something and something away&29.2&74.5\\
Move something up&34.3&49.5\\
Move something down&24.3&68.6\\
\bottomrule 
\end{tabular}
\label{categories}
\vspace{-0.3cm}
\end{table}

\mypara{Fine-grained action recognition results on SSv2.}
The paper's motivation is to improve the motion learning for contrastive learning. To further verify if \ourmethod indeed learns better motion feature, we extend the linear evaluation on the SSv2 dataset to a fine-grained split of video categories. As presented in Table~\ref{categories},  we compare \ourmethod with our base method, SimCLR~\cite{simclr}, on different categories, under the linear evaluation protocol. All the models are pre-trained on the training split of SSv2. \ourmethod and SimCLR get similar performances for the top categories, but \ourmethod significantly outperforms SimCLR on the bottom ones. Specifically, our method improves SimCLR by more than 40\% on ``Move something down''. Note that the bottom categories require accurate temporal features to be correctly recognized, indicating that \ourmethod does teach the model to learn motion features better. 

\begin{figure}[!t]
  \centering
  \includegraphics[width=0.85\linewidth]{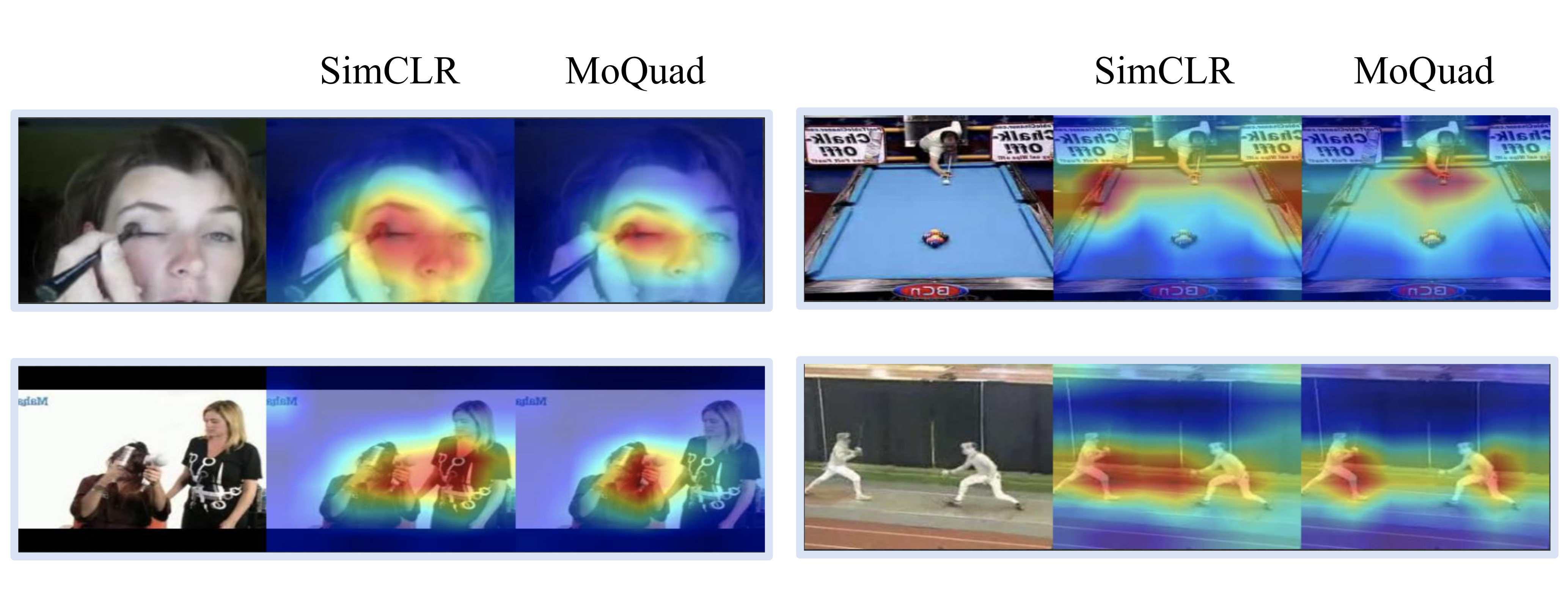}
  \vspace{-0.2cm}
   \caption{\textbf{Visualizing the region of interest (RoI).} We use CAM~\cite{cam} to plot the RoI of the model. \ourmethod focuses more on the moving regions than SimCLR.}
   \vspace{-0.5cm}
   \label{fig:heat}
\end{figure}
\mypara{RoI visualization.}
We visualize the Region of Interest (RoI) of \ourmethod and SimCLR in Fig.~\ref{fig:heat} using the tool provided by CAM \cite{cam}. Compared to SimCLR, \ourmethod better constrains the attention to the moving regions, providing a clue that our method does improve the learning of motion-oriented features.

\section{Conclusion}
This paper proposes a simple yet effective method (\ourmethod) to improve the learning of motion features in contrastive learning framework.
A carefully designed sample construction strategy is the core of \ourmethod, while two additional training strategies further improve the performance. By simply applying \ourmethod to SimCLR, extensive experiments show that we outperform state-of-the-art self-supervised video representation learning approaches on both action recognition and video retrieval.

\clearpage
%
%
\bibliographystyle{splncs04}
\bibliography{egbib}
\end{document}